\documentclass[11pt,letterpaper]{article}
\usepackage{emnlp2017}
\usepackage{times}
\usepackage{amsmath}
\usepackage{latexsym}
\usepackage{graphicx}
\usepackage{multirow}
\usepackage{tikz,tikz-qtree}

\newcommand{\LF}[1]{\ensuremath{\mathbf{#1}}}

\newcommand{\setof}[1]{\ensuremath{\left \{ #1 \right \}}}

\newcommand{\InferenceRule}[3]{
  \begin{tikzpicture}[level distance=36pt,sibling distance=12pt]
  \Tree [. \ensuremath{#1} \edge [->] node[auto=left]{\sc #2}; \ensuremath{#3} ]
  \end{tikzpicture}}
\newcommand{\InferenceRuleThree}[5]{
  \begin{tikzpicture}[level distance=56pt,sibling distance=12pt]
  \Tree [. \ensuremath{#1} \edge [->] node[auto=left]{\sc #2}; 
    [. \ensuremath{#3} \edge [->] node[auto=left]{\sc #4}; \ensuremath{#5} ] ]
  \end{tikzpicture}}
\newcommand{\SeqFormulas}[2]{
  \begin{tabular}{l}\ensuremath{#1}\\\ensuremath{#2}\end{tabular}}
\newcommand{\SeqFormulasThree}[3]{
  \begin{tabular}{l}\ensuremath{#1}\\\ensuremath{#2}\\\ensuremath{#3}\end{tabular}}
\newcommand{\SeqFormulasFour}[4]{
  \begin{tabular}{l}\ensuremath{#1}\\\ensuremath{#2}\\\ensuremath{#3}\\\ensuremath{#4}\end{tabular}}

\emnlpfinalcopy

\def\emnlppaperid{624}

\title{Determining Semantic Textual Similarity\\
using Natural Deduction Proofs}

\author{Hitomi Yanaka$^1$\\ {\tt hitomiyanaka@g.ecc.u-tokyo.ac.jp}\\
		\And
		\hspace{1.5cm}Koji Mineshima$^2$\\\hspace{1.5cm}{\tt mineshima.koji@ocha.ac.jp}\\
		\AND
		Pascual Mart\'{i}nez-G\'{o}mez$^3$ \\ {\tt pascual.mg@aist.go.jp}\\
		\And
		\hspace{1.5cm}Daisuke Bekki$^2$\\\hspace{1.5cm}{\tt bekki@is.ocha.ac.jp}\\
		\AND
        $^1${\rm The University of Tokyo} \\
        $^2${\rm Ochanomizu University} \\
        $^3${\rm Artificial Intelligence Research Center, AIST} \\
        {\rm Tokyo, Japan}
}

\date{}

\begin{document}

\maketitle

\begin{abstract}
Determining semantic textual similarity is a core research subject in natural language processing.
Since vector-based models for sentence representation often use shallow information, 
capturing accurate semantics is difficult.
By contrast, logical semantic representations capture deeper levels
of sentence semantics, but their symbolic nature does not offer graded
notions of textual similarity.
We propose a method for determining semantic textual similarity by combining
shallow features with features extracted from 
natural deduction proofs 
of bidirectional entailment relations between sentence pairs. 
For the natural deduction proofs, we use ccg2lambda, a higher-order automatic inference system, 
which converts 
Combinatory Categorial Grammar (CCG) derivation trees into semantic representations and 
conducts natural deduction proofs. Experiments show that our system 
was able to outperform other logic-based systems and that features derived from the 
proofs are effective for learning textual similarity.
\end{abstract}
\section{Introduction}

Determining semantic textual similarity (STS) is one of the most critical tasks in information retrieval and natural language processing. 
Vector-based sentence representation models have been widely used to compare and rank words, phrases or sentences using various similarity and relatedness scores~\cite{Find-similar,mitchell2010composition,DBLP:conf/icml/LeM14}.
Recently, neural network-based sentence representation models~\cite{MuellerAAAI2016, hill-cho-korhonen:2016:N16-1} have been proposed for learning textual similarity.
However, these vector-based models often use shallow information, such as words and characters, and 
whether they can account for phenomena such as negation and quantification is not clear.
Consider the sentences: \textit{Tom did not meet some of the players} and \textit{Tom did not meet any of the players}. 
If functional words such as \textit{some} or \textit{any} are ignored or represented as the same vector, then these sentences are to be represented by identical vectors.
However, the first sentence 
implies that there is a player who Tom did not meet, 
whereas the second sentence means that Tom 
did not meet anyone,  
so the sentences have different meanings.

Conversely, logic-based approaches 
have been successful in representing 
the meanings of complex sentences, having had a positive impact 
for applications such as recognizing textual entailment~\cite{D16-1242, mineshima2016building, abzianidze:2015:EMNLP, abzianidze:2016:*SEM}.
However, purely logic-based approaches only 
assess entailment or contradiction relations between sentences and do not offer graded notions of semantic similarity.

In this paper, we propose to leverage logic cues to learn textual similarity. Our hypothesis is that \emph{observing 
proof processes when testing the semantic relations is predictive of textual similarity}. We show that our approach 
can be more effective 
than systems that ignore these logic cues.

\section{Related Work}
Vector-based models of semantic composition have been widely studied with regards to calculating STS.
~\citet{mitchell-lapata:2008:ACLMain,mitchell2010composition} proposed a sentence vector model involving word vector addition or component-wise multiplication. Addition and multiplication
are commutative and associative and thus ignore word order. 
~\citet{polajnar-rimell-clark:2015:LSDSem} proposed a discourse-based sentence vector model considering extra-intra sentential context. Also, a categorical compositional distributional semantic model has been developed for recognizing textual entailment and for calculating STS~\cite{grefenstette-sadrzadeh:2011:EMNLP, kartsaklis-kalchbrenner-sadrzadeh:2014:P14-2, kartsaklis-sadrzadeh:2016:COLING}. However, these previous studies are mostly concerned with the structures of basic phrases or sentences
and do not address logical and functional words such as negations and connectives.
Neural network-based models of semantic composition~\cite{MuellerAAAI2016, hill-cho-korhonen:2016:N16-1} have also been proposed. 
Although these models achieve higher accuracy, their end-to-end nature
introduces challenges in the diagnosis of the reasons that make two sentences
to be similar or dissimilar to each other. These diagnosis capabilities may
play an important role in making the system explainable and also to guide
future system improvements in a more precise manner.
Our approach presented in this paper is partially inspired by the latter two objectives.

Meanwhile, some previous studies have proposed logic systems for capturing the semantic relatedness 
of sentences.
The Meaning Factory~\cite{bjerva:semeval14} 
uses both shallow and logic-based features for learning textual similarity. In this system, the overlap of predicates and entailment judgments are extracted as logic-based features.
UTexas~\cite{beltagy:semeval14} 
uses Probabilistic Soft Logic for learning textual similarity. In this system, each ground atom in the logical formulas has a probability based on distributional semantics of a word.
The weights of the logical formulas are calculated from the probabilities of their ground atoms and are extracted as features. These previous studies improved the accuracy by using logic-based features derived from 
the entailment results of first-order theorem proving in addition to using shallow features such as sentence lengths. 

In our study, we 
determine the semantic similarity of sentences based on the conception of proof-theoretic semantics~\cite{BekkiMineshima2016Luo}. 
The key idea is that not only the entailment results but also the \emph{theorem proving process} can be considered as features for learning textual similarity. That is, by taking into account not only whether a theorem is proved but also \textit{how} it is proved, we can capture the semantic relationships between sentence pairs in more depth.
 
Another difference between our study and previous logic systems is that we use higher-order predicate logic. Higher-order predicate logic is able to represent complex sentence semantics such as generalized quantifiers more precisely than first-order predicate logic. In addition, higher-order predicate logic makes the logical structure of a sentence more explicit than first-order predicate logic does, so it can simplify the process of proof search~\cite{miller-nadathur:1986:ACL}.

\section{System Overview}
Figure 1 shows 
an overview of the system which extracts features for learning textual similarity from 
logical proofs. To produce semantic representations of sentences and prove them automatically, we use ccg2lambda~\cite{martinezgomez-EtAl:2016:P16-4}, which is a semantic parser combined with an inference system based on natural deduction.  

First, sentences are parsed into syntactic trees based on Combinatory Categorial Grammar (CCG)~\cite{Steedman00}. CCG is a syntactic theory suitable for semantic composition 
from syntactic structures. Meaning representations are obtained based on semantic templates and combinatory rules 
for the CCG trees.
Semantic templates are defined manually based on formal semantics.
Combinatory rules specify the syntactic behaviors of words and compositional rules 
for the CCG trees.
In ccg2lambda, two wide-coverage CCG parsers, C\&C~\cite{clark2007wide} and EasyCCG~\cite{Lewis14a*ccg}, are used for converting tokenized sentences into CCG trees robustly. According to 
a previous study \cite{EACL2017}, EasyCCG 
achieves higher accuracy. Thus, 
when the output of both C\&C and EasyCCG can be proved, we use EasyCCG's output for creating features.

Second, the meanings of words 
are described 
using 
lambda terms. Semantic representations are obtained by combining lambda terms in accordance with the meaning composition rules specified in the CCG tree.
The semantic representations are based on Neo-Davidsonian event semantics~\cite{Parsons90,D16-1242}, in which every verb is decomposed into a predicate over events and a set of functional expressions relating the events.
Adverbs and prepositions are also represented as predicates over events.

\begin{figure}
\centerline{\includegraphics[bb=0.000000 0.000000 822.000000 274.000000,width=1.0\hsize]{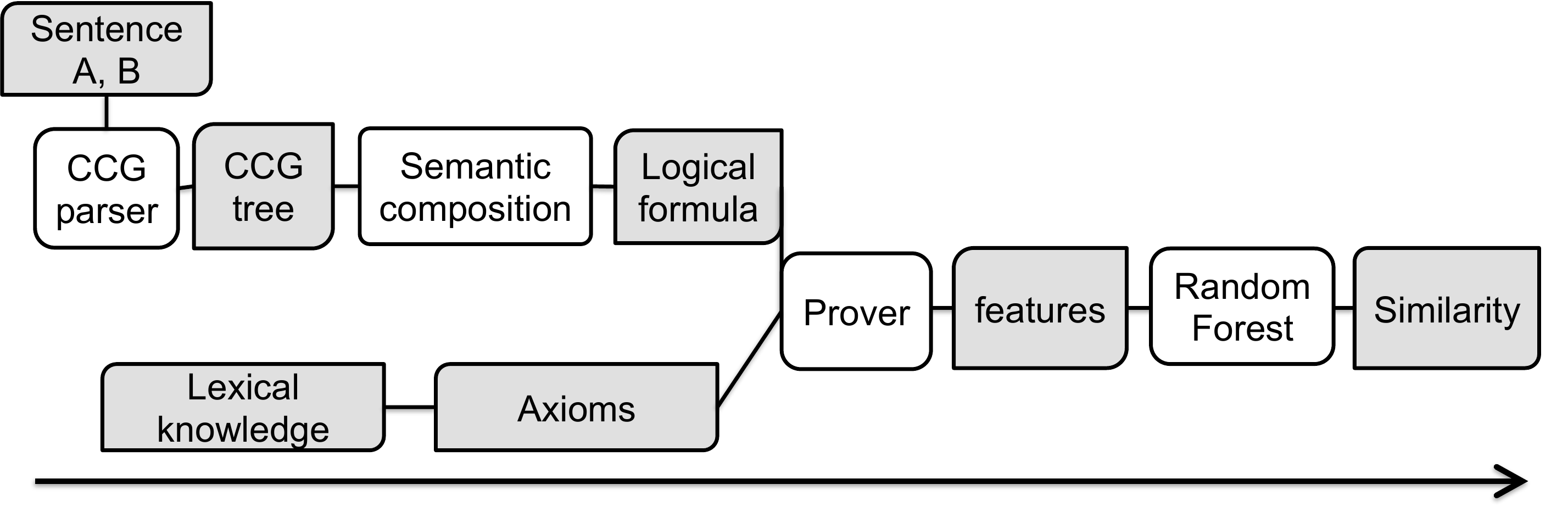}}
\caption{System overview.}
\label{Figure 1}
\end{figure}

Third, we attempt to prove entailment relations between sentence pairs.
For this purpose, we use Coq~\cite{opac-b1101046}, which can be used for efficient theorem-proving
for natural language inference using both first-order and higher-order logic~\cite{D16-1242}.
Coq's proof calculus is based on natural deduction~\cite{prawitz1965natural}, a proof system based on inference rules called introduction and elimination rules for logical connectives.
The inference system implemented in ccg2lambda 
using Coq 
achieves efficient automatic inference by feeding a set of predefined tactics and user-defined proof-search tactics to its interactive mode. 
The natural deduction system is particularly suitable for injecting external axioms during the theorem-proving process~\cite{EACL2017}.

Finally, features for learning textual similarity are extracted from the proofs produced by ccg2lambda during the theorem-proving process.
In this study, we
experimented with logistic regression, support vector regression and random forest regression, finding that random forest regression was the most effective. 
We therefore chose random forest regression for learning textual similarity, with its hyperparameters being optimized by grid search. The mean squared error (MSE) was used to measure the prediction performance of our system.

\section{Proof Strategy for Learning Textual Similarity}
\subsection{Overview of the proof strategy}
Sentence similarity depends on complex elements, 
such as word overlaps and semantic relations. 
We capture the similarity between the sentence pair $(A, B)$
as a function of the provability of bidirectional entailment relations for $(A, B)$ and combine it with shallow features. 
After obtaining logical formulas $A'$ and $B'$ from $A$ and $B$,
we attempt to prove the bidirectional entailment relations, $A' \Rightarrow B' $ and $B' \Rightarrow A'$. 
If the initial natural deduction proofs fail, we 
re-run the proof, adding relevant external axioms or skipping unproved sub-goals until the proof is 
completed.
After that, features for learning textual similarity are extracted by quantifying the provability of the bidirectional entailment relations.

The details of the procedure are as follows. First, we attempt a natural deduction proof
without using external axioms, aiming to prove entailment relations,
$A' \Rightarrow B'$ and $B' \Rightarrow A'$.
If both fail, then we check 
whether $A'$ contradicts $B'$, which amounts to proving
the negation of the original conclusion, namely
$A' \Rightarrow \neg B'$ and $B' \Rightarrow \neg A'$.

The similarity 
of a sentence pair tends to be higher when the negation of the conclusion can be proved, 
compared with the case 
where neither the conclusion nor its negation can be proved.
In the SICK (Sentences Involving Compositional Knowledge) dataset~\cite{MARELLI14.363} (see Section 6.1 for details),  70\% of the sentence pairs annotated as 
contradictory are assigned a relatedness score in [$3, 5$). 

Next, if we fail to prove entailment or contradiction,
that is, 
we cannot prove the conclusion or its negation, 
we identify an unproved sub-goal 
which is not matched by any predicate in the premise. 
We then attempt to prove $A' \Rightarrow B'$ and $B' \Rightarrow A'$ 
using axiom injection,
following the method introduced in \citet{EACL2017}.
In axiom injection, unproved sub-goals are candidates to form axioms. We focus only on 
predicates that share at least one argument with both the premise and the conclusion. 
This means that an axiom can be generated only if there is a predicate $p$ in the pool of 
premises and a predicate $q$ in a sub-goal and $p$ and $q$ share a variable in an argument position, possibly with the same case (e.g., Subject or Object). 

In generating axioms, the semantic relationships between the predicates in the premise and those in the conclusion are checked 
using lexical knowledge. In this study, we use WordNet~\cite{Miller:1995:WLD:219717.219748}
as the source of lexical knowledge. Linguistic relations between predicates
are checked in 
the following order: inflections, derivationally related forms, synonyms, antonyms, hypernyms, similarities, and hyponyms. If any one of these relations is found in the lexical knowledge, an axiom can be generated. Again, if the proof fails, we attempt to prove the negation of the conclusion 
using the axiom injection mechanism.

If the proof by axiom injection fails 
because of a lack of lexical knowledge, we obtain sentence similarity information from partial proofs by simply accepting the unproved sub-goals and 
forcibly completing the proof. After the proof is 
completed, information about the generated axioms and skipped sub-goals is used to create features.

\begin{figure}[t]
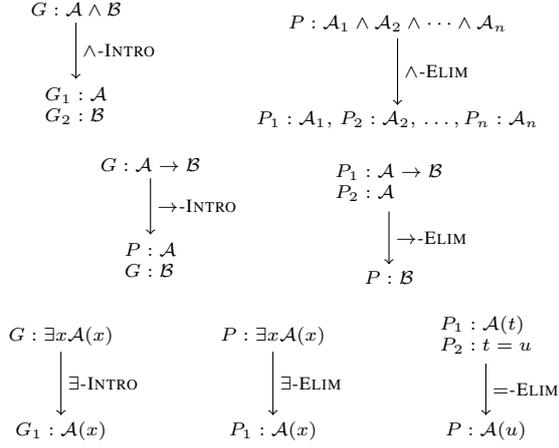

\scriptsize
\centering

\InferenceRule{G: \mathcal{A} \wedge \mathcal{B}}{$\wedge$-Intro}{
\SeqFormulas{G_1: \mathcal{A}}{G_2: \mathcal{B}}}
\hspace{4em}
\InferenceRule{P: \mathcal{A}_1 \wedge \mathcal{A}_2 \wedge \cdots \wedge \mathcal{A}_n}{$\wedge$-Elim}{P_1: \mathcal{A}_1, \, P_2:  \mathcal{A}_2, \, \ldots, P_n : \mathcal{A}_n}

\medskip

\InferenceRule{G: \mathcal{A} \to \mathcal{B}}{$\to$-Intro}{\SeqFormulas{P: \mathcal{A}}{G: \mathcal{B}}}
\hspace{3em}
\InferenceRule{\SeqFormulas{P_1: \mathcal{A} \to \mathcal{B}}{P_2: \mathcal{A}}}{$\to$-Elim}{P: \mathcal{B}}

\medskip

\InferenceRule{G: \exists x \mathcal{A}(x)}{$\exists$-Intro}{G_1: \mathcal{A}(x)}
\hspace{3em}
\InferenceRule{P: \exists x \mathcal{A}(x)}{$\exists$-Elim}{P_1: \mathcal{A}(x)}
\hspace{3em}
\InferenceRule{\SeqFormulas{P_1: \mathcal{A}(t)}{P_2: t = u}}{$=$-Elim}{P: \mathcal{A}(u)}

\caption{Example of the inference rules used in natural deduction.
$P, P_1, \ldots P_n$ are formulas in the premise,
while $G, G_1, G_2$ are formulas in the goal. 
The initial formulas are at the top, with the formulas obtained by applying the inference rules shown below.}
\label{InferenceRules}
\end{figure}

\subsection{Proving entailment relations}

As an illustration of how our natural deduction proof works,
consider the case of proving entailment 
for the following sentence pair:

\hspace{2em} $A$: A man is singing in a bar.

\hspace{2em} $B$: A man is singing.

The sentences $A$ and $B$ are mapped onto
logical formulas $A'$ and  $B'$ based on event semantics
via CCG-based semantic composition, as follows.
\begin{align*}
& \scalebox{0.9}{$A': \exists e_1 x_1 x_2(\LF{man}(x_1) \wedge \LF{sing}(e_1)
    \wedge (\LF{subj}(e_1) = x_1)$} \\
& \qquad \scalebox{0.9}{$\wedge \ \LF{bar}(x_2) \wedge \LF{in}(e_1, x_2))$} \\
& \scalebox{0.9}{$B': \exists e_1 x_1(\LF{man}(x_1) \wedge \LF{sing}(e_1)
 \wedge (\LF{subj}(e_1) = x_1))$}
\end{align*}
First, we attempt a natural deduction proof of $A' \Rightarrow B'$, setting $A'$ as the premise and $B'$ as the goal of the proof.
Then $A'$ and $B'$ are decomposed according to 
the inference rules.

Figure \ref{InferenceRules} shows the major inference rules we use in the proofs.
Inference rules in natural deduction are divided into
two types: introduction rules and elimination rules.
Introduction rules specify how to prove a formula in the goal, 
decomposing a goal formula into smaller sub-goals.
Elimination rules specify how to use a premise, decomposing a formula in the pool of premises into smaller ones.

\begin{figure}[t]
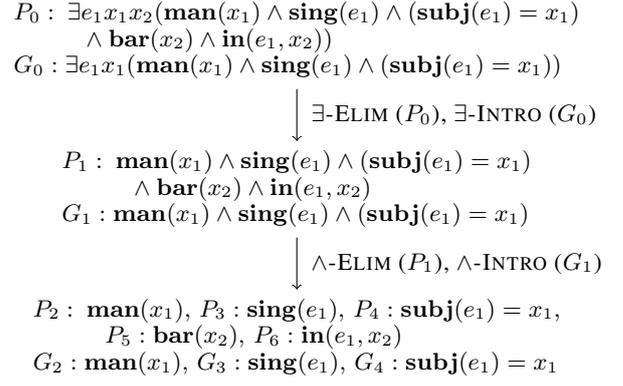

\footnotesize

\InferenceRuleThree{
\SeqFormulasThree{P_0: \ \exists e_1 x_1 x_2 (\LF{man}(x_1) \wedge \LF{sing}(e_1) \wedge (\LF{subj}(e_1) = x_1)}{
\hspace{3em} \wedge \ \LF{bar}(x_2) \wedge \LF{in}(e_1, x_2))}{
G_0: \exists e_1 x_1(\LF{man}(x_1) \wedge \LF{sing}(e_1) \wedge (\LF{subj}(e_1) = x_1))}}{
$\exists$-Elim ($P_0$), $\exists$-Intro ($G_0$)}{
\SeqFormulasThree{P_1: \ \LF{man}(x_1) \wedge \LF{sing}(e_1) \wedge (\LF{subj}(e_1) = x_1)}{
\hspace{3em} \wedge \ \LF{bar}(x_2) \wedge \LF{in}(e_1, x_2)}{
G_1: \LF{man}(x_1) \wedge \LF{sing}(e_1) \wedge (\LF{subj}(e_1) = x_1)}}{
$\wedge$-Elim ($P_1$), $\wedge$-Intro ($G_1$)}{
\SeqFormulasThree{P_2: \ \LF{man}(x_1), \, P_3: \LF{sing}(e_1), \,
P_4: \LF{subj}(e_1) = x_1,}{
\hspace{3em} P_5: \LF{bar}(x_2), \, P_6: \LF{in}(e_1, x_2)}{
G_2: \LF{man}(x_1), \, G_3: \LF{sing}(e_1), \, G_4: \LF{subj}(e_1) = x_1}}

\caption{The proof process for the example entailment relation.}
\label{ProcessEntail}
\end{figure}

The proof process 
for $A' \Rightarrow B'$ is shown in Figure \ref{ProcessEntail}.
Here $A'$ is initially set to the premise $P_0$ and $B'$ to the goal $G_0$.
$P_0$ and $G_0$ are then decomposed using
elimination rules ({\small$\wedge$-\textsc{Elim}, $\exists$-\textsc{Elim}})
and introduction rules ({\small$\wedge$-\textsc{Intro}, $\exists$-\textsc{Intro}}).
Then we obtain a set of premise formulas $\mathcal{P} = \setof{P_2, P_3, P_4, P_5, P_6}$,
and a set of sub-goals $\mathcal{G} = \setof{G_2, G_3, G_4}$.
The proof is performed by searching for a premise $P_i$
whose predicate and arguments match those of a given sub-goal $G_j$.
If such a logical premise is found, the sub-goal is removed.
In this example, the sub-goals $G_2$, $G_3$, and $G_4$
match the premises $P_2$, $P_3$, and $P_4$, respectively.
Thus, $A' \Rightarrow B'$ can be proved without introducing axioms.

Second, we attempt the proof in 
the opposite direction, $B' \Rightarrow A'$,
by switching $P_0$ and $G_0$ in Figure \ref{ProcessEntail}.
Again, by applying inference rules,
we obtain the following sets of premises $\mathcal{P}$ and sub-goals $\mathcal{G}$:

\smallskip
\small
\begin{tabular}{l}
$\mathcal{P} = \{P_2: \LF{man}(x_1), \, P_3: \LF{sing}(e_1),$ \\
 \hspace{2.4em} $P_4: \LF{subj}(e_1) = x_1\}$ \\
$\mathcal{G} = \{G_2: \LF{man}(x_1),\, G_3: \LF{sing}(e_1),$ \\
\hspace{2.4em} $G_4: \LF{subj}(e_1) = x_1,$ \\
\hspace{2.4em} $G_5: \LF{bar}(x_2), G_6: \LF{in}(e_1, x_2))\}$ \\
\end{tabular}
\normalsize

\noindent
Here, the two sub-goals $G_5$ and $G_6$ do not match any of the premises,
so the attempted proof of $B' \Rightarrow A'$ fails.
We therefore attempt to inject additional axioms, but in this case no predicate in $\mathcal{P}$
shares the argument $x_2$ of the predicates $\LF{bar}(x_2)$
and $\LF{in}(e_1,x_2)$ in $\mathcal{G}$.
Thus, no axiom can be generated. To obtain information from a partial proof, 
we forcibly complete the proof of $B' \Rightarrow A'$ by skipping the unproved sub-goals $\LF{bar}(x)$ and $\LF{in}(e_1,x_2)$.

\subsection{Proving the contradiction}

\begin{figure}[!t]
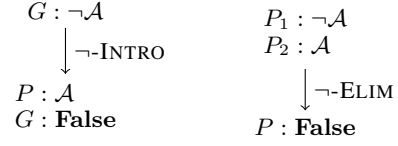

\centering
\footnotesize

\InferenceRule{G: \neg \mathcal{A}}{$\neg$-Intro}{\SeqFormulas{P: \mathcal{A}}{G: \LF{False}}}
\hspace{2em}
\InferenceRule{\SeqFormulas{P_1: \neg \mathcal{A}}{P_2: \mathcal{A}}}{$\neg$-Elim}{P: \LF{False}}

\caption{Inference rules of negation.}
\label{NegationRule}
\end{figure}

The proof strategy illustrated here
can be straightforwardly applied to proving the contradiction. 
In natural deduction, 
a negative formula of the form $\neg A$ can be defined as
$A \to \LF{False}$ (``the formula $A$ implies the contradiction''),
by using a propositional constant \LF{False} to encode the contradiction.
Thus, the inference rules for negation
can be taken as special cases of implication rules, as shown in Figure~\ref{NegationRule}.

As an illustration, let us consider the following sentence pair:

$A$: No man is singing.

$B$: There is a man singing loudly.

\noindent
Figure \ref{ProveContra} shows 
the proof process. The sentences
$A$ and $B$ are mapped to $P_0$ and $P_1$, respectively, via compositional semantics and
the goal $G_0$ is set to \LF{False}.
By decomposing $P_1$ using elimination rules
and then by combining $P_2, P_3$, and $P_4$,
we can obtain $P_6$.
From $P_0$ and $P_6$ we can then derive the contradiction.

These proofs are performed by an automated prover implemented on Coq, using tactics for first-order theorem proving.
When a proof is successful, 
Coq outputs the resulting proof (a proof term), from which
we can extract detailed information
such as the number of proof steps and 
the types of inference rules used.
In addition to the entailment/contradiction 
result,
information about 
the proof process is used to create 
features.

\section{Description of the Features}
To maximize accuracy when learning textual similarity, we adopt a hybrid approach 
that uses both logic-based features extracted from the natural deduction proof and other, non-logic-based features. 
All features are scaled to the [$0, 1$] range.

\subsection{Logic-based Features}
We propose 15 features consisting of nine different types of logic-based features. Six of these feature types are derived from the bidirectional natural deduction proofs: six features are extracted from the direct proof ($A' \Rightarrow B'$) and another six from the reverse proof ($B' \Rightarrow A'$). The remaining three feature types are derived from semantic representations of the sentence pairs. The feature types are as follows.

\begin{figure}[t]
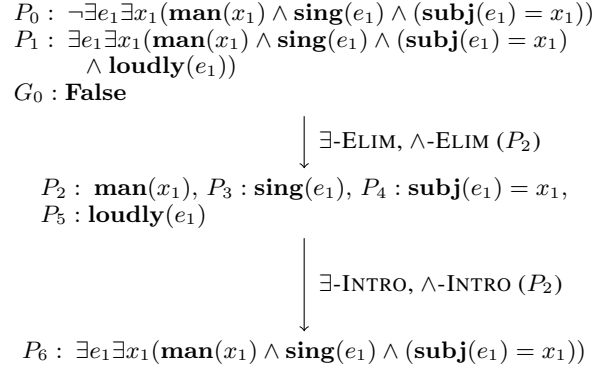

\footnotesize

\InferenceRuleThree{\SeqFormulasFour{
P_0: \ \neg \exists e_1 \exists x_1(\LF{man}(x_1) \wedge \LF{sing}(e_1) \wedge (\LF{subj}(e_1) = x_1))}{
P_1: \ \exists e_1 \exists x_1(\LF{man}(x_1) \wedge \LF{sing}(e_1) \wedge (\LF{subj}(e_1) = x_1)}{
\hspace{3em} \wedge \ \LF{loudly}(e_1))}{
G_0: \LF{False}}}{
$\exists$-Elim, $\wedge$-Elim ($P_2$)
}{
\SeqFormulas{
P_2: \ \LF{man}(x_1), \, P_3: \LF{sing}(e_1), \, P_4: \LF{subj}(e_1) = x_1,}{
P_5: \LF{loudly}(e_1)}}{
$\exists$-Intro, $\wedge$-Intro ($P_2$)}{
P_6: \ \exists e_1 \exists x_1 (\LF{man}(x_1) \wedge \LF{sing}(e_1) \wedge (\LF{subj}(e_1) = x_1))}

\caption{\label{ProveContra}Proof process for the contradiction example.}
\vspace{-0.3cm}
\end{figure}

\noindent{\bf Logical inference result.}
As stated in Section 4, we 
include features to distinguish the case where either the conclusion or its negation can be proved from the 
one where neither can be proved. If the conclusion can be proved, the feature is set to 1.0. If the negation of the conclusion can be proved, the feature is set to 0.5. If neither can be proved, the feature is set to 0.0.

\noindent{\bf Axiom probabilities.}
The probability of an axiom and the number of axioms appearing in the proof are used 
to create features. The probability of an axiom is defined as the inverse of the length of the shortest path that connects the senses in the is-a (hypernym/hyponym) taxonomy in WordNet. When multiple axioms are used in the proof, the average of the probabilities of the axioms 
is extracted as a feature. If the proof 
can be completed without using axioms, the feature is set to 1.0.

\noindent{\bf Proved sub-goals.}
Given that 
proofs can be obtained either by proving all the sub-goals or skipping unproved sub-goals, we use the proportion of 
proved sub-goals as a feature. Our assumption is that if there are more unproved sub-goals then the sentence pair is less similar.
When there are $m$ logical formulas in the premise pool and $n$ proved sub-goals, we set the feature to $n/m$. If the theorem can be proved without skipping any sub-goals, the feature is set to 1.0. 
It may be the case that the number of sub-goals is so large that some sub-goals remain unproved even after axiom injection. 
Since the proportion of unproved sub-goals is decreased by axiom injection, 
we use the proportion of unproved sub-goals both with and without axiom injection as features.

\noindent{\bf Cases in unproved sub-goals.}
Subject or object words 
can affect the similarity of sentence pairs. Therefore,
the number of each case in unproved sub-goals, 
like $\LF{subj}(e_1)$ in Figures \ref{ProcessEntail} and \ref{ProveContra}, is used as a feature. 
Here, we count subjective, objective, and dative cases.

\noindent{\bf Proof steps.}
In general, complex theorems are difficult to prove and in such cases the sentence pairs
are considered to be less similar. We therefore use the number of Coq's proof steps, namely the number of inference rule applications in a given proof, as a feature.

\noindent{\bf Inference rules.}
The complexity of a natural deduction proof can be measured 
in terms of the inference rules used 
for each proof step. We therefore extract the relative frequency with which each inference rule is used in the proof as a feature. We check seven inference rules for natural deduction using Coq (cf.  Figure \ref{InferenceRules}):
introduction and elimination rules for conjunction
({\small$\wedge$-\textsc{Intro}, $\wedge$-\textsc{Elim}}),
 implication ({\small$\to$-\textsc{Intro}, $\to$-\textsc{Elim}}),
and existential quantification ({\small$\exists$-\textsc{Intro}, $\exists$-\textsc{Elim}}),
and the elimination rule for equality ({\small$=$-\textsc{Elim}}).

\noindent{\bf Predicate overlap.}
Intuitively, the more predicates overlap between the premise and the conclusion, the more likely it is that the inference can be proved. We therefore use the proportion of predicates that overlap between the premise and the conclusion as a feature.

\noindent{\bf Semantic type overlap.}
Each semantic representation in higher-order logic has a semantic type, such as \textsf{Entity} for entities and \textsf{Prop} for propositions. As with predicates, we use the degree of semantic type overlap between the premise and the conclusion as a feature.

\noindent{\bf Existence of negative clauses.}
Whether or not the premise or conclusion contain negative clauses is an effective measure of similarity. In semantic representations, negative clauses are represented by the negation operator $\neg$, so we check for negation operators in the premise and the conclusion and set this feature to 1.0 if either contains one. 

\begin{table*}[!t]
\begin{center}
\scalebox{0.97}{
\begin{tabular}{ccccc} \hline
   ID & Sentence1 & Sentence2 & Entailment & Score\\ \hline \hline
   23 & There is no biker jumping in the air. & A lone biker is jumping in the air & \textit{no} & 4.2 \\ \hline
   1412 & Men are sawing logs. & Men are cutting wood.  & \textit{yes} & 4.5 \\ \hline
   9963 & The animal is grazing on the grass. & The cop is sitting on a police bike. & \textit{unknown} & 1 \\ \hline 
\end{tabular}
}
\vspace{-0.3cm}
\caption{
\label{tab:examples} Examples in the SICK dataset with different entailment labels and similarity scores.}
\vspace{-0.5cm}
\end{center}
\end{table*}

\subsection{Non-logic-based Features}
We also use the following eight non-logic-based features.

\noindent{\bf Noun/verb overlap.}
We extract and lemmatize all nouns and verbs from the sentence pairs and use the degrees of overlap of the noun and verb lemmas as features.

\noindent{\bf Part-of-speech overlap.}
We obtain part-of-speech (POS) tags for all words in the sentence pairs by first tokenizing them with the Penn Treebank Project tokenizer\footnote{ftp://ftp.cis.upenn.edu/pub/treebank/public\_html/\\
tokenization.html}
and then POS tagging them with C\&C POS tagger~\cite{curran2003investigating}.
The degree of overlap between the sentences' POS tags is used as a feature.

\noindent{\bf Synset overlap.}
For each sentence in the pair, we obtain the set containing all the synonym lemmas (the synset) for the words in the sentence.
The degree of overlap between the sentences' synsets is used as a feature.

\noindent{\bf Synset distance.}
For 
each word in the first sentence, we compute the maximum path similarity between its synset and the synset of any other word in the second sentence. Then, we use the average of maximum path similarities as a feature.

\noindent{\bf Sentence length.}
If the conclusion sentence is long,  
there will possibly be many sub-goals in the proof. We therefore use the average of the sentence lengths and the difference in length between the premise and the conclusion sentences as features.

\noindent{\bf String similarity.}
We use the similarity of the sequence of characters within the sentence pairs as a feature. 
The Python {\sl Difflib}\footnote{https://docs.python.org/3.5/library/difflib.html} function returns the similarity between two sequences as a floating-point value in 
[$0, 1$]. This measure is given by $2.0*M / T$, where $T$ is the total number of elements in both sequences and $M$ is the number of matches. This feature is 1.0 if the sequences are identical and 0.0 if they have nothing in common.

\noindent{\bf Sentence similarity from vector space models.}
We calculate sentence similarity by using three major vector space models, TF-IDF, latent semantic analysis (LSA)~\cite{LSA}, and latent Dirichlet allocation (LDA)~\cite{LDA}. We use these cosine similarities as features.

\noindent{\bf Existence of passive clauses.}
Passive clauses have an influence on similarity. In CCG trees, passive clauses are represented 
using the syntactic category $S_{pss} \backslash NP$. We check for the occurrence of passive clauses in the premise and conclusion, and if either of them contains a passive clause then the feature is set to 1.0.

\section{Experiments and Evaluation}
\subsection{Experimental Conditions}
We evaluated our system\footnote{Available at https://github.com/mynlp/ccg2lambda.} using two datasets: the SemEval-2014 version of the SICK dataset~\cite{MARELLI14.363} and the SemEval-2012 version of the MSR-paraphrase video corpus dataset (MSR-vid)~\cite{semeval2012}.
The experimental conditions were as follows.

\begin{table}[!t]
\begin{center}
\scalebox{0.93}{
\begin{tabular}{lccc} \hline
                          & $\gamma$ & $\rho$  & MSE     \\ \hline \hline
   Mueller et al. (2016)  & $0.882$  & $0.835$ & $0.229$ \\ \hline \hline 
   Our system             & $0.838$  & $0.796$ & $0.561$ \\  \hline
   SemEval2014 Best Score & $0.828$  & $0.769$ & $0.325$ \\ \hline
   The Meaning Factory    & $0.827$  & $0.772$ & $0.322$ \\  \hline
   UTexas                 & $0.714$  & $0.674$ & $0.499$ \\ \hline
   Baseline               & $0.653$  & $0.745$ & $0.808$ \\  \hline
\end{tabular}
}
\vspace{-0.3cm}
\caption{\label{tab:results_sick} Results on the test split of SICK dataset.}
\vspace{-0.5cm}
\end{center}
\end{table}

\subsubsection{The SICK dataset}
The SICK dataset is a dataset for studying 
STS as well as for recognizing textual entailment (RTE). It was originally
developed for evaluating compositional distributional semantics, so it contains
logically challenging expressions such as quantifiers, negations, conjunctions
and disjunctions. The dataset contains $9927$ sentence pairs with a $5000$/$4927$
training/test split. These sentence pairs are manually
annotated with three types of labels 
\textit{yes} (entailment), \textit{no} (contradiction), or
\textit{unknown} (neutral) as well as a semantic relatedness scores in 
[$1, 5$] (see Table~\ref{tab:examples} for a sample).

In this dataset, sentence pairs whose gold entailment labels are 
\textit{no} tend to be scored a little more highly than the average, whereas 
those whose labels are  
\textit{unknown} have a wide range of scores. Thus, we set the baseline of the relatedness score to 5 when the gold entailment label was \textit{yes} and to 3 when the label was \textit{no} or \textit{unknown}.

We compared our system 
with the following systems: the state-of-the-art neural network-based system~\cite{MuellerAAAI2016}; the best system~\cite{zhao:semeval14} from SemEval-2014; and two of the logic-based systems stated in Section 2: namely The Meaning Factory~\cite{bjerva:semeval14} and UTexas~\cite{beltagy:semeval14}. The Pearson correlation coefficient $\gamma$, Spearman's rank correlation coefficient $\rho$, and the MSE were used as the evaluation metrics.

\medskip

\subsubsection{The MSR-vid dataset}
The MSR-vid dataset is our second dataset for the STS task and contains $1500$ sentence pairs with a $750$/$750$ training/test split. 
All sentence pairs are annotated with semantic relatedness scores in the range [0, 5].
We used this dataset to compare our system with the best system from SemEval-2012~\cite{bar:semeval12} and the logic-based UTexas system~\cite{beltagy:acl14}. We used the Pearson correlation coefficient $\gamma$ as the evaluation metric.

\subsection{Results}

Table~\ref{tab:results_sick} shows the results of our 
experiments with the SICK dataset. Although the state-of-the-art neural network-based system yielded the best results overall, our system achieved higher scores than SemEval-2014 submissions, including the two logic-based systems (The Meaning Factory and UTexas), in terms of Pearson correlation and Spearman's correlation. 

The main reason for our system's lower performance in terms of MSE is that some theorems could not be proved 
because of a lack of lexical knowledge. In the current work,
we only consider word-level knowledge (word-for-word paraphrasing); we may expand the knowledge base in the future by using more external resources.

As we mentioned above, the sentence pairs annotated as \textit{unknown} produced a wide range of scores. The Pearson correlation of the \textit{unknown} portion of the SICK dataset was 0.766, which suggests that our logic-based system can also be applied to neutral sentence pairs.

\begin{table}[!t]
\begin{center}
\begin{tabular}{lc} \hline
                          & $\gamma$ \\ \hline \hline
   SemEval2012 Best Score & $0.873$  \\ \hline \hline 
   Our system             & $0.853$  \\ \hline 
   Beltagy et al. (2014)  & $0.830$  \\ \hline
\end{tabular}
\vspace{-0.3cm}
\caption{\label{tab:results_msrvid} Results on the test split of MSR-vid.}
\vspace{-0.3cm}
\end{center}
\end{table}

\begin{table}[!htbp]
\begin{center}
\scalebox{0.89}{
\begin{tabular}{lccc} \hline
                             & $\gamma$          & $\rho$            & MSE              \\ \hline \hline
   Predicate overlap         & $\mathbf{0.691}$  & $0.609$           & $\mathbf{0.734}$ \\ \hline
   Inference rules           & $0.632$           & $\mathbf{0.619}$  & $0.794$          \\  \hline
   Probability of axioms     & $0.543$           & $0.540$           & $0.865$          \\ \hline
   Proof steps               & $0.458$           & $0.494$           & $0.915$          \\ \hline
   Proved sub-goals          & $0.432$           & $0.443$           & $0.926$          \\  \hline
   Logical inference result  & $0.386$           & $0.399$           & $0.939$          \\ \hline
   Unproved sub-goals' case  & $0.301$           & $0.307$           & $0.973$          \\  \hline
   Semantic type overlap     & $0.245$           & $0.219$           & $0.987$          \\ \hline 
   Negative clauses          & $0.163$           & $0.323$           & $1.004$          \\ \hline \hline
   Noun/verb overlap         & $0.661$           & $0.554$           & $0.763$          \\ \hline
   Vector space model        & $0.594$           & $0.510$           & $0.857$          \\ \hline
   String similarity         & $0.414$           & $0.418$           & $0.977$          \\ \hline
   Synset overlap            & $0.382$           & $0.341$           & $0.978$          \\ \hline
   Synset distance           & $0.352$           & $0.330$           & $0.999$          \\ \hline
   Part-of-speech overlap    & $0.349$           & $0.346$           & $0.954$          \\ \hline
   Sentence length           & $0.231$           & $0.240$           & $0.993$          \\ \hline 
   Passive clauses           & $0.023$           & $0.046$           & $1.017$          \\ \hline \hline
   Only logic-based          & $0.798$           & $0.760$           & $0.613$          \\ \hline 
   Only non logic-based      & $0.793$           & $0.732$           & $0.621$          \\ \hline \hline
   All                       & $\mathbf{0.838}$  & $\mathbf{0.796}$  & $\mathbf{0.561}$ \\ \hline
\end{tabular}
}
\caption{\label{tab:results_feats} Results when training our regressor with each feature group in isolation.}
\end{center}
\end{table}
Table~\ref{tab:results_msrvid} shows the results of our 
experiments with the MSR-vid dataset. These results also indicate that our logic-based system achieved higher accuracy than the other logic-based systems.

\begin{table*}
\begin{center}
\small
\begin{tabular}{rlcccc} \hline
                      &                                                                       &                        & Pred                 & Pred                  &                          \\
ID                    & Sentence Pair                                                         & Gold                   & +logic               & -logic                & Entailment               \\ \hline \hline
\multirow{2}{*}{642}  & A person is climbing a rock with a rope, which is pink.               & \multirow{2}{*}{5.0}   & \multirow{2}{*}{4.9} & \multirow{2}{*}{4.1}  & \multirow{2}{*}{Yes}     \\
                      & A rock is being climbed by a person with a rope, which is pink.       &                        &                      &                       &                          \\ \hline
\multirow{2}{*}{1360} & The machine is shaving the end of a pencil.                           & \multirow{2}{*}{4.7}   & \multirow{2}{*}{4.6} & \multirow{2}{*}{3.8}  & \multirow{2}{*}{Yes}     \\
                      & A pencil is being shaved by the machine.                              &                        &                      &                       &                          \\ \hline
\multirow{2}{*}{891}  & There is no one on the shore.                                         & \multirow{2}{*}{3.6}   & \multirow{2}{*}{3.7} & \multirow{2}{*}{2.6}  & \multirow{2}{*}{No}      \\
                      & A bunch of people is on the shore.                                    &                        &                      &                       &                          \\ \hline
\multirow{2}{*}{1158} & A woman is removing ingredients from a bowl.                          & \multirow{2}{*}{3.3}   & \multirow{2}{*}{3.5} & \multirow{2}{*}{4.1}  & \multirow{2}{*}{No}      \\
                      & A woman is adding ingredients to a bowl.                              &                        &                      &                       &                          \\ \hline
\multirow{2}{*}{59}   & Kids in red shirts are playing in the leaves.                         & \multirow{2}{*}{3.9}   & \multirow{2}{*}{3.8} & \multirow{2}{*}{3.1}  & \multirow{2}{*}{Unknown} \\
                      & Three kids are jumping in the leaves.                                 &                        &                      &                       &                          \\ \hline
\multirow{2}{*}{71}   & There is no child lying in the snow and making snow angels.           & \multirow{2}{*}{3.3}   & \multirow{2}{*}{3.3} & \multirow{2}{*}{4.1}  & \multirow{2}{*}{Unknown} \\
                      & Two people in snowsuits are lying in the snow and making snow angels. &                        &                      &                       &                          \\ \hline
\end{tabular}
\caption{\label{tab:examples_pos} Examples for which our regressor trained only with logic-based features performs better than when using non-logic features. ``Gold'': correct score, ``Pred+logic'': prediction score only with logic-based features, ``Pred-logic'': prediction score only with non-logic-based features.}
\end{center}
\end{table*}

Table~\ref{tab:results_feats} shows evaluation results for each feature group in isolation, 
showing that inference rules and predicate overlaps are the most effective features. 
Compared with the non-logic-based features, the logic-based features achieved a slightly higher accuracy, a point that will be analyzed in more detail in the next section. 
Overall, our results show that combining logic-based features with non logic-based ones is an effective method for determining textual similarity.

\subsection{Positive examples and error analysis}

\begin{table*}[t]
\begin{center}
\small
\begin{tabular}{rlccl} \hline
ID&Sentence Pair&Gold&System&Axiom \\ \hline \hline
   \multirow{2}{*}{3974} & A girl is awakening.          & \multirow{2}{*}{4.9} & \multirow{2}{*}{3.6} & $\forall x (\LF{awaken}(x) \rightarrow \LF{wake}(x))$ \\
                         & A girl is waking up.          &                      &                      & $\forall x (\LF{awaken}(x) \rightarrow \LF{up}(x))$   \\ \hline

   \multirow{2}{*}{4833} & A girl is filing her nails.   & \multirow{2}{*}{4.2} & \multirow{2}{*}{1.8} & $\forall x (\LF{nail}(x) \rightarrow \LF{manicure}(x))$ \\
                         & A girl is doing a manicure.   &                      &                      & $\forall x (\LF{file}(x) \rightarrow \LF{do}(x))$       \\ \hline

   \multirow{3}{*}{1941} & \multirow{3}{*}{\begin{tabular}{@{}l@{}}A woman is putting the baby into a trash can. \\ A person is putting meat into a skillet. \end{tabular}} & \multirow{3}{*}{1.0} & \multirow{3}{*}{3.3} & $\forall x (\LF{woman}(x) \rightarrow \LF{person}(x))$   \\
                         &                                                                                                                                                  &                      &                      & $\forall x (\LF{trash}(x) \rightarrow \LF{skillet}(x))$  \\
                         &                                                                                                                                                  &                      &                      & $\forall x (\LF{baby}(x) \rightarrow \LF{meat}(x))$      \\ \hline
\end{tabular}
\caption{\label{tab:examples_neg} Error examples when training the regressor only with logic-based features.}
\end{center}
\end{table*}

Table~\ref{tab:examples_pos} shows some 
examples for which the prediction score was better when using
logic-based features than when using non-logic-based ones.

For IDs 642 and 1360,
one sentence contains a passive clause while the other sentence does not. In such cases, the sentence pairs are not superficially similar. 
By using logical formulas based on event semantics we were able to interpret the sentence containing the passive clause correctly and judge that the passive and non-passive sentences are similar to each other.

In ID 891, 
one sentence contains a negative clause 
while the other does not. 
Using shallow features, the word overlap is small and the prediction score was much lower than the correct score. Our logic-based method, however, interpreted the first sentence as a negative existential formula
of the form $\neg \exists x \mathcal{P}(x)$
and the second sentence as an existential formula $\exists x \mathcal{P'}(x)$.
Thus, it could easily handle the semantic difference between the positive and negative sentences.

In ID 1158, by contrast, the proportion of word overlap is so high that the prediction score with non-logic-based features was much higher than the correct score. Our method, however, was able to 
prove the contradiction using an antonym axiom of the form
$\forall x (\LF{remove}(x) \rightarrow \neg \LF{add}(x))$ from WordNet and thus predict the score correctly.

In ID 59, the proportion of word overlap is low, so the prediction score with non-logic-based features was lower than the correct score. Our method, however, was able to 
prove the partial entailment relations
for the sentence pair and  
thus predict the score correctly.
Here the logic-based method captured the common meaning of the sentence pair:
both sentences talk about the kids playing in the leaves.

Finally, in ID 71, the prediction score with non-logic-based features was much higher than the correct score. There are two reasons for this phenomenon: 
negations tend to be omitted in non-logic-based features such as TF-IDF 
and the proportion of word overlap is high. However, as logical formulas and 
proofs can handle negative clauses correctly, 
our method was able to predict the score correctly.

Table~\ref{tab:examples_neg} shows 
examples where using only logic-based features produced erroneous results. In ID 3974, the probability of axiom $\forall x (\LF{awaken}(x) \!\rightarrow\! \LF{up}(x))$ was low (0.25) and thus the prediction score was lower than the correct score. 
Likewise, in ID 4833, the probability of axiom
$\forall x (\LF{file}(x) \!\rightarrow\! \LF{do}(x))$ was very low (0.09) and thus the prediction score was negatively affected.
In these cases, we need to consider phrase-level axioms such as $\forall x (\LF{awaken}(x) \!\rightarrow\! \LF{wake}\_\LF{up}(x))$
and $\forall x (\LF{file}\_\LF{nail}(x) \!\rightarrow\! \LF{do}\_\LF{manicure}(x))$ using a paraphrase database. This, however, is an issue for future study. In ID 1941, the system wrongly proved the bidirectional entailment relations by adding external axioms, so the prediction score was much higher than the correct score. 
Setting the threshold for the probability of an axiom may be an effective way of improving our axiom-injection method.

\section{Conclusion}
We have developed a hybrid method for learning textual similarity by combining features based on logical proofs of bidirectional entailment relations with non-logic-based features. The results of our experiments on two datasets show that our system was able to outperform other logic-based systems. In addition, the results show that information about the natural deduction proof process can be used to create effective features for learning textual similarity. Since these logic-based features provide accuracy improvements that are 
largely additive with those provided by non-logic-based features, neural network-based systems may also benefit from using them.

In future work, we will refine our system so that it can be applied to other tasks such as question answering. Compared with neural network-based systems, our natural deduction-based system can not only assess how similar sentence pairs are, but also explain what the sources of similarity/dissimilarity are by referring to information about sub-goals in the proof. Given 
this interpretative ability, we believe that our logic-based system 
may also be of benefit to other natural language processing tasks, such as question answering and text summarization.

\section*{Acknowledgments}
We thank the three anonymous reviewers for their detailed comments. This work was supported by JST CREST Grant Number JPMJCR1301, Japan.

\bibliography{emnlp2017}

\begin{thebibliography}{35}
\expandafter\ifx\csname natexlab\endcsname\relax\def\natexlab#1{#1}\fi

\bibitem[{Abzianidze(2015)}]{abzianidze:2015:EMNLP}
Lasha Abzianidze. 2015.
\newblock A tableau prover for natural logic and language.
\newblock In \emph{Proceedings of the 2015 Conference on Empirical Methods in
  Natural Language Processing (EMNLP-15)}, pages 2492--2502, Lisbon, Portugal.
  Association for Computational Linguistics.

\bibitem[{Abzianidze(2016)}]{abzianidze:2016:*SEM}
Lasha Abzianidze. 2016.
\newblock Natural solution to {F}ra{C}a{S} entailment problems.
\newblock In \emph{Proceedings of the 5th Joint Conference on Lexical and
  Computational Semantics}, pages 64--74, Berlin, Germany. Association for
  Computational Linguistics.

\bibitem[{Agirre et~al.(2012)Agirre, Cer, Diab, and
  Gonzalez-Agirre}]{semeval2012}
Eneko Agirre, Daniel Cer, Mona Diab, and Aitor Gonzalez-Agirre. 2012.
\newblock Sem{E}val-2012 {T}ask 6: A pilot on semantic textual similarity.
\newblock In \emph{Proceedings of the 6th International Workshop on Semantic
  Evaluation (SemEval-2012)}, pages 385--393, Montr\'{e}al, Canada. Association
  for Computational Linguistics.

\bibitem[{B\"{a}r et~al.(2012)B\"{a}r, Biemann, Gurevych, and
  Zesch}]{bar:semeval12}
Daniel B\"{a}r, Chris Biemann, Iryna Gurevych, and Torsten Zesch. 2012.
\newblock {UKP}: Computing semantic textual similarity by combining multiple
  content similarity measures.
\newblock In \emph{Proceedings of the Sixth International Workshop on Semantic
  Evaluation {(SemEval-2012)}}, pages 435--440, Montr\'{e}al, Canada.
  Association for Computational Linguistics.

\bibitem[{Bekki and Mineshima(2017)}]{BekkiMineshima2016Luo}
Daisuke Bekki and Koji Mineshima. 2017.
\newblock Context-passing and underspecification in dependent type semantics.
\newblock In Stergios Chatzikyriakidis and Zhaohui Luo, editors, \emph{Modern
  Perspectives in Type Theoretical Semantics}, Studies of Linguistics and
  Philosophy, pages 11--41. Springer.

\bibitem[{Beltagy et~al.(2014{\natexlab{a}})Beltagy, Erk, and
  Mooney}]{beltagy:acl14}
Islam Beltagy, Katrin Erk, and Raymond Mooney. 2014{\natexlab{a}}.
\newblock Probabilistic soft logic for semantic textual similarity.
\newblock In \emph{Proceedings of the 52nd Annual Meeting of the Association
  for Computational Linguistics (ACL-2014)}, pages 1210--1219, Baltimore,
  Maryland. Association for Computational Linguistics.

\bibitem[{Beltagy et~al.(2014{\natexlab{b}})Beltagy, Roller, Boleda, Erk, and
  Mooney}]{beltagy:semeval14}
Islam Beltagy, Stephen Roller, Gemma Boleda, Katrin Erk, and Raymond Mooney.
  2014{\natexlab{b}}.
\newblock {UT}exas: Natural language semantics using distributional semantics
  and probabilistic logic.
\newblock In \emph{Proceedings of the 8th International Workshop on Semantic
  Evaluation (SemEval-2014)}, pages 796--801, Dublin, Ireland. Association for
  Computational Linguistics and Dublin City University.

\bibitem[{Bertot and Castran(2010)}]{opac-b1101046}
Yves Bertot and Pierre Castran. 2010.
\newblock \emph{Interactive Theorem Proving and Program Development: Coq'Art
  The Calculus of Inductive Constructions}.
\newblock Springer Publishing Company, Incorporated, New York, USA.

\bibitem[{Bjerva et~al.(2014)Bjerva, Bos, van~der Goot, and
  Nissim}]{bjerva:semeval14}
Johannes Bjerva, Johan Bos, Rob van~der Goot, and Malvina Nissim. 2014.
\newblock The {M}eaning {F}actory: Formal semantics for recognizing textual
  entailment and determining semantic similarity.
\newblock In \emph{Proceedings of the 8th International Workshop on Semantic
  Evaluation (SemEval-2014)}, pages 642--646, Dublin, Ireland. Association for
  Computational Linguistics and Dublin City University.

\bibitem[{Blei et~al.(2003)Blei, Ng, and Jordan}]{LDA}
David~M. Blei, Andrew~Y. Ng, and Michael~I. Jordan. 2003.
\newblock Latent dirichlet allocation.
\newblock \emph{Journal of Machine Learning}, 3:993--1022.

\bibitem[{Clark and Curran(2007)}]{clark2007wide}
Stephen Clark and James~R. Curran. 2007.
\newblock Wide-coverage efficient statistical parsing with {CCG} and log-linear
  models.
\newblock \emph{Computational Linguistics}, 33(4):493--552.

\bibitem[{Curran and Clark(2003)}]{curran2003investigating}
James~R Curran and Stephen Clark. 2003.
\newblock Investigating {GIS} and smoothing for maximum entropy taggers.
\newblock In \emph{Proceedings of the tenth conference on European chapter of
  the Association for Computational Linguistics-Volume 1}, pages 91--98.
  Association for Computational Linguistics.

\bibitem[{Deerwester et~al.(1990)Deerwester, Dumais, Landauer, and
  Harshman}]{LSA}
Scott Deerwester, Susan~T. Dumais, Thomas~K. Landauer, and Richard Harshman.
  1990.
\newblock Indexing by latent semantic analysis.
\newblock \emph{Journal of the American Society for Information Science},
  41(6):391--407.

\bibitem[{Grefenstette and Sadrzadeh(2011)}]{grefenstette-sadrzadeh:2011:EMNLP}
Edward Grefenstette and Mehrnoosh Sadrzadeh. 2011.
\newblock Experimental support for a categorical compositional distributional
  model of meaning.
\newblock In \emph{Proceedings of the 2011 Conference on Empirical Methods in
  Natural Language Processing (EMNLP-2011)}, pages 1394--1404, Edinburgh,
  Scotland, UK. Association for Computational Linguistics.

\bibitem[{Hill et~al.(2016)Hill, Cho, and
  Korhonen}]{hill-cho-korhonen:2016:N16-1}
Felix Hill, Kyunghyun Cho, and Anna Korhonen. 2016.
\newblock Learning distributed representations of sentences from unlabelled
  data.
\newblock In \emph{Proceedings of the 2016 Conference of the North American
  Chapter of the Association for Computational Linguistics: Human Language
  Technologies}, pages 1367--1377, San Diego, California. Association for
  Computational Linguistics.

\bibitem[{Kartsaklis et~al.(2014)Kartsaklis, Kalchbrenner, and
  Sadrzadeh}]{kartsaklis-kalchbrenner-sadrzadeh:2014:P14-2}
Dimitri Kartsaklis, Nal Kalchbrenner, and Mehrnoosh Sadrzadeh. 2014.
\newblock Resolving lexical ambiguity in tensor regression models of meaning.
\newblock In \emph{Proceedings of the 52nd Annual Meeting of the Association
  for Computational Linguistics (ACL-2014)}, pages 212--217, Baltimore,
  Maryland. Association for Computational Linguistics.

\bibitem[{Kartsaklis and Sadrzadeh(2016)}]{kartsaklis-sadrzadeh:2016:COLING}
Dimitri Kartsaklis and Mehrnoosh Sadrzadeh. 2016.
\newblock Distributional inclusion hypothesis for tensor-based composition.
\newblock In \emph{Proceedings of the 26th International Conference on
  Computational Linguistics: Technical Papers (COLING-2016)}, pages 2849--2860,
  Osaka, Japan. The COLING 2016 Organizing Committee.

\bibitem[{Le and Mikolov(2014)}]{DBLP:conf/icml/LeM14}
Quoc~V. Le and Tomas Mikolov. 2014.
\newblock Distributed representations of sentences and documents.
\newblock In \emph{Proceedings of the 31th International Conference on Machine
  Learning, ({ICML}-2014)}, pages 1188--1196, Beijing, China.

\bibitem[{Lewis and Steedman(2014)}]{Lewis14a*ccg}
Mike Lewis and Mark Steedman. 2014.
\newblock {A* CCG} parsing with a supertag-factored model.
\newblock In \emph{Proceedings of the 2014 Conference on Empirical Methods in
  Natural Language Processing (EMNLP-2014)}, pages 990--1000, Doha, Qatar.
  Association for Computational Linguistics.

\bibitem[{Marelli et~al.(2014)Marelli, Menini, Baroni, Bentivogli, Bernardi,
  and Zamparelli}]{MARELLI14.363}
Marco Marelli, Stefano Menini, Marco Baroni, Luisa Bentivogli, Raffaella
  Bernardi, and Roberto Zamparelli. 2014.
\newblock A {SICK} cure for the evaluation of compositional distributional
  semantic models.
\newblock In \emph{Proceedings of the 9th International Conference on Language
  Resources and Evaluation (LREC-2014)}, pages 216--223, Reykjavik, Iceland.
  European Language Resources Association.

\bibitem[{Mart\'{i}nez-G\'{o}mez et~al.(2016)Mart\'{i}nez-G\'{o}mez, Mineshima,
  Miyao, and Bekki}]{martinezgomez-EtAl:2016:P16-4}
Pascual Mart\'{i}nez-G\'{o}mez, Koji Mineshima, Yusuke Miyao, and Daisuke
  Bekki. 2016.
\newblock ccg2lambda: A compositional semantics system.
\newblock In \emph{Proceedings of ACL-2016 System Demonstrations}, pages
  85--90, Berlin, Germany. Association for Computational Linguistics.

\bibitem[{Mart\'{i}nez-G\'{o}mez et~al.(2017)Mart\'{i}nez-G\'{o}mez, Mineshima,
  Miyao, and Bekki}]{EACL2017}
Pascual Mart\'{i}nez-G\'{o}mez, Koji Mineshima, Yusuke Miyao, and Daisuke
  Bekki. 2017.
\newblock On-demand injection of lexical knowledge for recognising textual
  entailment.
\newblock In \emph{Proceedings of the 15th Conference of the European Chapter
  of the Association for Computational Linguistics (EACL-2017)}, pages
  710--720, Valencia, Spain. Association for Computational Linguistics.

\bibitem[{Miller and Nadathur(1986)}]{miller-nadathur:1986:ACL}
Dale~A. Miller and Gopalan Nadathur. 1986.
\newblock Some uses of higher-order logic in computational linguistics.
\newblock In \emph{Proceedings of the 24th Annual Meeting of the Association
  for Computational Linguistics}, pages 247--256, New York, New York, USA.
  Association for Computational Linguistics.

\bibitem[{Miller(1995)}]{Miller:1995:WLD:219717.219748}
George~A. Miller. 1995.
\newblock Word{N}et: A lexical database for {E}nglish.
\newblock \emph{Communications of the ACM}, 38(11):39--41.

\bibitem[{Mineshima et~al.(2015)Mineshima, Mart\'{i}nez-G\'{o}mez, Miyao, and
  Bekki}]{D16-1242}
Koji Mineshima, Pascual Mart\'{i}nez-G\'{o}mez, Yusuke Miyao, and Daisuke
  Bekki. 2015.
\newblock Higher-order logical inference with compositional semantics.
\newblock In \emph{Proceedings of the 2015 Conference on Empirical Methods in
  Natural Language Processing (EMNLP-2015)}, pages 2055--2061, Lisbon,
  Portugal. Association for Computational Linguistics.

\bibitem[{Mineshima et~al.(2016)Mineshima, Tanaka, Mart{\'\i}nez-G{\'o}mez,
  Miyao, and Bekki}]{mineshima2016building}
Koji Mineshima, Ribeka Tanaka, Pascual Mart{\'\i}nez-G{\'o}mez, Yusuke Miyao,
  and Daisuke Bekki. 2016.
\newblock Building compositional semantics and higher-order inference system
  for a wide-coverage {J}apanese {CCG} parser.
\newblock In \emph{Proceedings of the 2016 Conference on Empirical Methods in
  Natural Language Processing}, pages 2236--2242, Austin, Texas. Association
  for Computational Linguistics.

\bibitem[{Mitchell and Lapata(2008)}]{mitchell-lapata:2008:ACLMain}
Jeff Mitchell and Mirella Lapata. 2008.
\newblock Vector-based models of semantic composition.
\newblock In \emph{Proceedings of the 46th Annual Meeting of the Association
  for Computational Linguistics (ACL-08)}, pages 236--244, Columbus, Ohio.
  Association for Computational Linguistics.

\bibitem[{Mitchell and Lapata(2010)}]{mitchell2010composition}
Jeff Mitchell and Mirella Lapata. 2010.
\newblock {Composition in distributional models of semantics}.
\newblock \emph{Cognitive Science}, 34(8):1388--1429.

\bibitem[{Mueller and Thyagarajan(2016)}]{MuellerAAAI2016}
Jonas Mueller and Aditya Thyagarajan. 2016.
\newblock Siamese recurrent architectures for learning sentence similarity.
\newblock In \emph{Proceedings of the 30th {AAAI} Conference on Artificial
  Intelligence (AAAI-2016)}, pages 2786--2792, Arizona, {USA}. Association for
  the Advancement of Artificial Intelligence.

\bibitem[{Parsons(1990)}]{Parsons90}
Terence Parsons. 1990.
\newblock \emph{{E}vents in The Semantics of {English}: a Study in Subatomic
  Semantics}.
\newblock {MIT} {P}ress, Cambridge, USA.

\bibitem[{Polajnar et~al.(2015)Polajnar, Rimell, and
  Clark}]{polajnar-rimell-clark:2015:LSDSem}
Tamara Polajnar, Laura Rimell, and Stephen Clark. 2015.
\newblock An exploration of discourse-based sentence spaces for compositional
  distributional semantics.
\newblock In \emph{Proceedings of the 1st Workshop on Linking Computational
  Models of Lexical, Sentential and Discourse-level Semantics}, pages 1--11,
  Lisbon, Portugal. Association for Computational Linguistics.

\bibitem[{Prawitz(1965)}]{prawitz1965natural}
Dag Prawitz. 1965.
\newblock \emph{Natural Deduction -- A Proof-Theoretical Study}.
\newblock Almqvist \& Wiksell, Stockholm, Sweden.

\bibitem[{Steedman(2000)}]{Steedman00}
Mark Steedman. 2000.
\newblock \emph{{T}he Syntactic Process}.
\newblock {MIT} {P}ress, Cambridge, USA.

\bibitem[{Wong and Raghavan(1984)}]{Find-similar}
S.~K.~M. Wong and Vijay~V. Raghavan. 1984.
\newblock Vector space model of information retrieval: A reevaluation.
\newblock In \emph{Proceedings of the 7th Annual International ACM SIGIR
  Conference on Research and Development in Information Retrieval}, pages
  167--185.

\bibitem[{Zhao et~al.(2014)Zhao, Zhu, and Lan}]{zhao:semeval14}
Jiang Zhao, Tiantian Zhu, and Man Lan. 2014.
\newblock {ECNU}: One stone two birds: Ensemble of heterogenous measures for
  semantic relatedness and textual entailment.
\newblock In \emph{Proceedings of the 8th International Workshop on Semantic
  Evaluation (SemEval-2014)}, pages 271--277, Dublin, Ireland. Association for
  Computational Linguistics and Dublin City University.

\end{thebibliography}
\bibliographystyle{emnlp_natbib}

\end{document}